\documentclass[11pt]{article}

\usepackage{amsmath, amssymb}
\usepackage{array}
\usepackage{graphicx}
\usepackage{booktabs}
\usepackage{multirow}
\usepackage{url}
\usepackage{hyperref}
\usepackage{enumitem}
\usepackage{geometry}
\usepackage{tabularx}
\usepackage{abstract}
\usepackage{tikz}
\usepackage[table]{xcolor}
\usepackage{arydshln}
\usepackage{eso-pic}
\usetikzlibrary{positioning,arrows.meta,fit,calc}

\newcommand{\rot}[1]{\rotatebox{90}{\scriptsize #1}}
\newcolumntype{Y}{>{\centering\arraybackslash}X}

\title{
Visual Product Search Benchmark
}

\author{
Karthik Sulthanpete Govindappa \\
nyris GmbH \\
Berlin, Germany \\
}

\date{}

\begin{document}

\maketitle

\begin{abstract}
\noindent Reliable product identification from images is a critical requirement in industrial and commercial applications, particularly in maintenance, procurement, and operational workflows where incorrect matches can lead to costly downstream failures. At the core of such systems lies the visual search component, which must retrieve and rank the exact object instance from large and continuously evolving catalogs under diverse imaging conditions. This report presents a structured benchmark of modern visual embedding models for instance-level image retrieval, with a focus on industrial applications. A curated set of open-source foundation embedding models, proprietary multi-modal embedding systems, and domain-specific vision-only models are evaluated under a unified image-to-image retrieval protocol. The benchmark includes curated datasets, which includes industrial datasets derived from production deployments in Manufacturing, Automotive, DIY, and Retail, as well as established public benchmarks. Evaluation is conducted without post-processing, isolating the retrieval capability of each model. The results provide insight into how well contemporary foundation and unified embedding models transfer to fine-grained instance retrieval tasks, and how they compare to models explicitly trained for industrial applications. By emphasizing realistic constraints, heterogeneous image conditions, and exact instance matching requirements, this benchmark aims to inform both practitioners and researchers about the strengths and limitations of current visual embedding approaches in production-level product identification systems. An interactive companion website presenting the benchmark results, evaluation details, and additional visualizations is available at \url{https://benchmark.nyris.io}.
\end{abstract}

\section{Introduction}

Product identification from images is a central capability of the systems developed at nyris. Over the past decade, we have partnered with businesses across a wide range of sectors, including retail, e-commerce, manufacturing, industrial automation, heavy machinery, and the automotive industry. These collaborations have provided extensive firsthand insight into the structure, scale, and challenges of image-based object identification under production constraints.
\newline

\noindent The product identification system developed at nyris consists of three primary components:
\begin{enumerate}[label=(\roman*)]
    \item a pre-processing engine that assesses the quality, suitability of input images, and localizes salient objects for downstream processing,
    \item a visual search engine that operates over a large-scale image database to generate a ranked list of candidate matches,
    \item a post-processing engine that applies additional filtering and decision logic to predict the final match.
\end{enumerate}

\noindent Among these components, the visual search engine is the most critical. Without high-quality retrieval and ranking performance at this stage, downstream identification becomes unreliable or may fail entirely. As a result, image retrieval forms the foundation of the overall system.
\newline

\noindent This report focuses exclusively on the visual search component of image-based product identification. The decision to narrow the scope in this way is motivated by several initiatives. 
\newline

\noindent First, systematic benchmarking and continuous improvement of visual search performance, especially for product identification, are essential to meeting the reliability expectations of customers operating production systems. While such evaluations have historically been conducted internally, we share this study to increase transparency for both existing and prospective customers of our technology, and to provide the research community with practical insight into the behavior of modern embedding models in instance-level image retrieval settings.
\newline

\noindent Second, we aim to foster closer collaboration between industry and the research community in the area of image-based product identification for industrial applications. The delivery of such solutions in maintenance workflows of manufacturing, heavy machinery, energy, and the automotive sector is extremely challenging. These challenges arise from limited labeled data, data regulations, domain gap in data distribution, extreme visual similarity between components, diverse imaging conditions, and the requirement for deep domain expertise to identify objects. In such settings, recommending visually similar alternatives is insufficient, the system must precisely identify the exact item of interest. We believe that progress in addressing the underlying challenges and open questions could be accelerated through structured collaborations between academia and industry.
\newline

\noindent With this motivation, we conduct a structured comparative evaluation of modern embedding models to better understand their behavior in practical instance-level retrieval scenarios. The study evaluates a curated selection of open-source models, proprietary foundation models, and specialist models developed at nyris, all assessed in a consistent image-to-image retrieval setting.

\section{Related Work}

In this section, we review prior work on instance-level image retrieval, embedding-based and metric learning approaches for visual search, and the evolution of foundation and vision-language models used for large-scale retrieval. We also discuss existing benchmark datasets that have shaped evaluation practices in instance-level retrieval. Rather than providing an exhaustive survey, we focus on work most relevant to practical, large-scale visual search systems and to the industrial instance identification scenarios considered in this study.

\subsection{Instance-Level Image Retrieval}

At nyris, we formulate visual search as an instance-level image retrieval problem. Instance-level image retrieval aims to retrieve images depicting the same physical object instance as the query image, rather than objects belonging to the same semantic category. Early work by Sivic and Zisserman introduced this formulation by framing object matching as a retrieval problem using local feature representations and bag-of-visual-words indexing\cite{sivic2003video}. Subsequent work by Philbin et al. demonstrated that instance retrieval could be scaled to large datasets using inverted file indexing and spatial verification, establishing many of the principles that continue to underpin modern retrieval systems \cite{philbin2007object}.
\newline

\noindent We adopt an information retrieval formulation rather than a classification-based approach due to fundamental challenges related to data scale, distribution, and system scalability. In practice, the product catalogs we operate on often contain thousands to millions of distinct products, exhibiting extreme class imbalance and highly variable numbers of images per product. Moreover, a set of images related to certain product frequently include non-representative or auxiliary images, such as brand logos, close-up detail shots, or packaging visuals. Product catalogs are also inherently dynamic, which means, new products are continuously introduced, existing products are removed, and associated images are frequently added, updated, or replaced. In several domains, catalogs additionally contain hundreds or even thousands of visually similar items, where discriminative differences are subtle and may involve fine-grained geometric details, surface textures, or markings.
\newline

\noindent Under these conditions, classification-based approaches become infeasible both computationally and statistically. Closed-set classifiers require frequent retraining to accommodate catalog changes, suffer from severe class imbalance, and cannot naturally handle unseen product instances. As noted in prior work, when the number of object instances is very large and continuously evolving, instance recognition is more appropriately cast as an image retrieval problem rather than classification \cite{tan2021instance}. Furthermore, embedding-based and metric learning approaches focus on modeling similarity relationships and have been shown to generalize more effectively in open-set scenarios where new classes are introduced at test time, whereas traditional softmax-based classifiers tend to overfit to training classes and perform poorly on novel instances \cite{liu2020few}.
\newline

\noindent Consequently, we formulate visual product search using learned embedding representations rather than closed-set classification, which naturally supports open-set recognition and incremental catalog updates without requiring frequent model retraining.

\subsection{Metric Learning and Embedding Models}

The emergence of deep representation learning in the mid-2010s marked a turning point for instance-level image retrieval. Early work by Hadsell et al.~\cite{hadsell2006dimensionality} demonstrated that convolutional neural networks could be trained end-to-end to learn invariant similarity metrics, establishing the foundation for modern deep metric learning. This paradigm was further advanced by the introduction of the triplet loss in FaceNet~\cite{schroff2015facenet}, which showed that embedding spaces could be explicitly optimized to preserve relative similarity relationships between instances at scale.
\newline

\noindent Subsequent research unified these approaches under the broader framework of deep metric learning, emphasizing the importance of globally structured and discriminative embedding spaces rather than isolated pairwise or triplet constraints~\cite{musgrave2020metric}. These advances established embedding-based retrieval as a scalable and effective alternative to classification-based recognition, particularly in open-set scenarios where the number of object instances is large and continuously evolving.
\newline

\noindent Embedding models have since become a core component of modern large-scale information retrieval systems. By mapping images, text, or other modalities into shared vector spaces, they enable efficient nearest-neighbor search at scale and support incremental catalog updates without retraining. As a result, embedding-based retrieval now underpins a wide range of applications, including large-scale visual search \cite{gordo2016deep}, recommendation systems \cite{covington2016deep}, and Retrieval-Augmented Generation (RAG) applications \cite{lewis2020retrieval} and Agentic AI workflows \cite{yao2022react}.
\newline

\noindent In computer vision, the introduction of large-scale foundation models further expanded the role of embeddings as general-purpose visual representations. Vision--language models such as CLIP~\cite{radford2021learning}, ALIGN~\cite{jia2021scaling}, and CoCa~\cite{yu2022coca}, trained on web-scale image--text pairs, demonstrated strong zero-shot generalization across diverse visual and multimodal tasks. These models highlighted the transferability of learned embedding spaces beyond their original training objectives.
\newline

\noindent Building on these developments, recent work has adapted vision-language models into unified embedding systems explicitly optimized for retrieval-centric applications. Works such as VLM2Vec~\cite{jiang2024vlm2vec}, Jina Embeddings v4~\cite{gunther2025jina}, and Nomic Embed Vision~\cite{nussbaum2024nomic} aim to produce embeddings that generalize across modalities, domains, and retrieval tasks without task-specific fine-tuning. While these models are often designed for semantic and multi-modal retrieval, their effectiveness for fine-grained, instance-level image retrieval under realistic industrial constraints remains an open question, which this benchmark seeks to systematically evaluate.

\subsection{Benchmarks for Instance-Level Image Retrieval}

The progress of instance-level image retrieval has been closely tied to the availability of well-defined benchmark datasets and evaluation protocols. Early benchmarks such as the Oxford and Paris landmark datasets, later revisited as ROxford and RParis, established standardized evaluation procedures and enabled systematic comparison of retrieval methods across varying levels of visual difficulty~\cite{radenovic2018revisiting}. These datasets emphasized fine-grained visual matching under changes in viewpoint, illumination, and partial occlusion, and remain widely used for evaluating instance-level retrieval models.
\newline

\noindent Subsequent large-scale benchmarks further expanded evaluation toward realistic and diverse retrieval scenarios. The Google Landmarks Dataset v2 (GLDv2) introduced a substantially larger and more challenging benchmark for instance-level recognition and retrieval, featuring significant visual diversity and real-world noise~\cite{weyand2020google}. In parallel, the Stanford Online Products (SOP) dataset became a standard benchmark for deep metric learning, focusing on fine-grained product instance discrimination with limited images per instance~\cite{oh2016deep}. These benchmarks have played a central role in advancing representation learning and retrieval methods in both academic and applied settings.
\newline

\noindent More recent benchmark initiatives aim to evaluate embedding models at scale and across broader retrieval conditions. The Universal Image Embeddings (UnED) dataset was introduced to assess the generalization of image embeddings across diverse content and retrieval tasks~\cite{ypsilantis2023towards}. Similarly, the ILIAS benchmark focuses explicitly on large-scale instance-level image retrieval with heterogeneous, distractor-rich galleries, reflecting realistic deployment scenarios where relevant instances must be retrieved from noisy and diverse search spaces~\cite{kordopatis2025ilias}.
\newline

\noindent Despite their importance, publicly available benchmarks often underrepresent the constraints and failure modes encountered in practical product identification systems. Real-world applications frequently involve niche domains characterized by extreme visual similarity between instances, strong domain shift between catalog and query images, limited or weakly descriptive metadata, and continuously evolving product catalogs. Such challenges are not often reflected in standard benchmark datasets.
\newline

\noindent Observing this gap, our benchmark complements existing efforts by evaluating modern visual embedding models on a combination of established public benchmarks and industrial datasets derived from practical deployments. Rather than replacing existing benchmarks, the intention of our work is to provide additional insight into the behavior of contemporary embedding representations when applied to instance-level image retrieval tasks.

\section{Datasets}
\label{sec:datasets}

To evaluate the instance-level image retrieval capabilities of the curated set of embedding models, we selected eight datasets spanning both industrial and publicly available domains. Four datasets originate from nyris’ internal deployments, covering manufacturing, automotive, DIY, and retail scenarios, while the remaining four are established public benchmarks commonly used to evaluate fine-grained image retrieval models. Together, these datasets span a broad spectrum of domain difficulty, intra-class similarity, image quality, and query conditions, enabling a comprehensive assessment of retrieval performance across both real-world and research-oriented settings.

\begin{table}[htbp]
\centering
\footnotesize
\setlength{\tabcolsep}{4pt}
\renewcommand{\arraystretch}{1.05}
\newcolumntype{Y}{>{\raggedleft\arraybackslash}X}

\begin{tabularx}{\linewidth}{p{4mm} X c c Y Y Y Y}

\toprule

& \textbf{Dataset} & \textbf{\shortstack{Retrieval\\Setup}} & \textbf{\shortstack{Gallery\\Type}} & \textbf{\shortstack{\#Images\\(Query)}} & \textbf{\shortstack{\#Images\\(Gallery)}} & \textbf{\shortstack{\#Products\\(Query)}} & \textbf{\shortstack{\#Products\\(Gallery)}} \\

\midrule

\multirow{4}{*}{\rotatebox{90}{\textbf{Private}}}
& Clips-and-Connectors v1   & Inter & Closed & 3,624 & 200,496  & 453 & 12,531 \\
& Furniture v1   & Inter & Closed & 371  & 85,165  & 347 & 50,641 \\
& DIY v1         & Inter & Closed & 5,008 & 221,376  & 5,008 & 221,376 \\
& Automotive v1  & Inter & Closed & 2,062 & 19,610 & 379 & 10,379 \\

\midrule

\multirow{3}{*}{\rotatebox{90}{\textbf{Public}}}
& SOP & Intra & Closed & 60,502 & 60,502  & 11,316 & 11,316 \\
& ILIAS & Inter & Open & 1,232  & 504,715 & 2,500  & 1,000$^{\dagger}$ \\
& Products-10K & Intra & Closed & 11,200 & 11,200  & 3,450  & 3,450 \\

\bottomrule

\end{tabularx}

\caption{Overview of the datasets used for evaluating instance-level image retrieval. 
For intra-retrieval datasets, the same split is used as query and gallery, with self-matches excluded during evaluation. $^{\dagger}$In the ILIAS dataset, the gallery contains 1,000 unique products, while the remaining images correspond to distractors that do not belong to any query product.}

\label{tab:datasets_overview}
\end{table}

\subsection{Dataset Structure}

To ensure consistency across datasets with differing structures and evaluation protocols, we introduce unified terminology to describe dataset splits, retrieval protocols, and gallery composition.\\

\noindent We distinguish between two retrieval setups. \textit{Intra-Retrieval} refers to an intra-split setting in which a single dataset split serves simultaneously as the query set and the gallery. Each image in the split is treated as a query against the entire set, and the query image itself is excluded from the ranked results during evaluation. In contrast, \textit{Inter-Retrieval} refers to a query--gallery setting in which queries and gallery images are drawn from distinct splits. This setup more closely reflects real-world production systems, where queries originate from user-captured imagery and the gallery consists of curated reference images.\\

\noindent We further characterize datasets based on the gallery product distributions. \textit{Closed-set} galleries contain only images corresponding to valid product or object instances relevant to the retrieval task. \textit{Open-set} galleries, in contrast, include a substantial number of images, such as distractors \cite{philbin2008lost}, that do not correspond to any query instance. This configuration simulates large-scale retrieval scenarios in which relevant instances must be identified within a broad and potentially noisy search space.

\subsection{Internal Benchmark Datasets}

The internal datasets are derived from production systems deployed by nyris, a company specializing in image-based product identification for industrial and commercial applications. The  visual search solutions provided by nyris are used in maintenance, procurement, and operational workflows, where accurate instance-level identification is critical and catalog updates are frequent. These datasets are designed to reflect real-world production constraints, including strong domain shift between catalog and query images, high intra-class similarity, and heterogeneous capture conditions.

\paragraph{Clips-and-Connectors v1.}
\label{dataset:fastenersv1}

This dataset represents one of the most challenging industrial benchmarks in our evaluation. It is derived from a client who is a global manufacturer of fastening and assembly solutions for the automotive and industrial sectors, where precise identification of small mechanical components is critical for maintenance and supply chain operations. The dataset contains 12,531 unique fasteners, including screws, bolts, nuts, washers, and pins, many of which differ only in minute geometric or dimensional properties. The gallery consists of 200,496 CAD-rendered images, generated from 3D models with 16 viewpoints per instance using a controlled multi-camera setup. These images provide clean, canonical representations. The query set contains 3,624 real-world images covering 453 fasteners, captured under uncontrolled industrial conditions with variations in lighting, background, viewpoint, occlusion, and reflections. This dataset evaluates a model’s ability to perform fine-grained instance retrieval and generalize from synthetic reference imagery to real user-captured input, a common requirement in industrial maintenance workflows.

\paragraph{Furniture v1.}
\label{dataset:furniturev1}
This dataset reflects a retail-oriented product identification scenario that involves visually rich consumer goods. The client from which this dataset was derived operates one of the world’s largest furniture catalogs, where product identification must bridge the gap between professionally produced catalog imagery and user-captured photos. The dataset contains 50,641 unique products. The gallery includes 85,165 high-quality catalog images, either professionally photographed or generated during product design. The query set contains 371 user-captured images representing 347 products, curated to reflect realistic search conditions with significant variation in viewpoint, lighting, and background complexity.

\paragraph{DIY v1.}
\label{dataset:diyv1}
This dataset is prepared from the data catalog of a client who is a large European DIY and home improvement retailer offering a broad assortment of tools, hardware, and building materials. This represents a large-scale retail catalog with substantial visual diversity across product categories. The gallery consists of 221,376 catalog images representing 127,597 products, predominantly captured against clean, uniform backgrounds. The query set includes 5,008 user-captured images, each corresponding to a distinct product. These images exhibit substantial variability in background clutter, lighting, orientation, and scene context.

\paragraph{Automotive v1.}
\label{dataset:automotivev1}
This evaluation dataset was curated with the support of a major European distributor of automotive spare parts that serves professional workshops and service centers. Accurate identification of vehicle components from user-captured images is critical for their repair and maintenance workflows. The gallery images were gathered from the distributor's vast catalog of 2M+ products, and the compatible query images were curated from the pool of request images processed by nyris Visual Search API. The gallery consists of 19{,}610 professionally produced catalog images representing 17{,}965 distinct products. These images were originally created for e-commerce and technical documentation purposes. The query set comprises 2{,}062 user-captured images representing 379 products, photographed in garages, workshops, and service environments. This dataset evaluates fine-grained instance-level retrieval under real-world capture conditions typical of automotive maintenance scenarios.

\subsection{Public Benchmark Datasets}
The public benchmarks provide scale, diversity, and comparability with existing instance-level image retrieval literature and were originally introduced to support systematic evaluation of fine-grained visual recognition and retrieval methods.

\paragraph{Stanford Online Products (SOP).}
\label{dataset:sop}
The Stanford Online Products (SOP) dataset was introduced to study deep metric learning for instance-level product retrieval and has become a standard benchmark in the field \cite{oh2016deep}. It contains 120,053 images representing 22,634 product instances across 12 categories collected from e-commerce listings. Each product instance is represented by only a small number of images, enforcing fine-grained discrimination.
In this study, we evaluate models using the test split only under an intra-retrieval protocol, where each image acts as a query against the remaining images of the same split, excluding the query image itself.

\paragraph{Products 10K.}
\label{dataset:products_10k}
Products-10K was originally introduced for fine-grained visual recognition using a classification-based evaluation protocol, with explicit training and test splits \cite{bai2020products}. The dataset contains over 214,000 images spanning more than 10,000 product categories, captured under diverse indoor and outdoor conditions. In this study, we do not follow the classification protocol. Instead, we evaluate models using only the test split under an intra-retrieval setup analogous to SOP. This choice allows us to assess instance-level retrieval performance in a zero-shot setting and ensures consistency across public benchmarks.
Although Products-10K could alternatively be evaluated using the training split as a reference gallery and the test split as queries, we do not adopt this setting in order to avoid introducing assumptions about supervised reference construction and to maintain a unified retrieval protocol across datasets.

\paragraph{ILIAS.}
\label{dataset:ilias}
ILIAS is a recent large-scale benchmark explicitly designed for instance-level image retrieval at scale, with a focus on realistic retrieval conditions and distractor-rich galleries \cite{kordopatis2025ilias}. It follows an inter-retrieval setup with a heterogeneous gallery containing a substantial number of non-relevant images. This configuration reflects realistic large-scale retrieval scenarios in which relevant instances must be retrieved from a gallery containing a large number of unrelated images.

\section{Models}

We evaluate a curated set of models suitable for image-to-image instance retrieval (see Table~\ref{tab:models_overview}). All models produce fixed-dimensional image embeddings that can be directly used for nearest-neighbor search. They differ in architectural design and scope, ranging from general-purpose foundation models to domain-specific industrial retrieval systems.\\

\noindent We organize the evaluated models along two dimensions: architecture and capability. Architecturally, we distinguish between Bi-Encoder models, Unified Embedding models, and Vision-Only models. In terms of capability, models are categorized as either generic or domain-specific. Generic models are typically trained at web scale and designed for broad applicability, including multi-modal semantic search and cross-domain retrieval tasks. In contrast, domain-specific models are optimized for particular industrial or instance-level retrieval scenarios.\\

\noindent Unless stated otherwise, all models are evaluated in an image-to-image retrieval setting, using only their image embedding pathways and without any modifications to native pre-processing or model configuration.

\begin{table}[htbp]
\centering
{\scriptsize
\setlength{\tabcolsep}{4pt}
\renewcommand{\arraystretch}{1.05}
\begin{tabularx}{\linewidth}{l X X X X r}
\toprule
\textbf{Model} & \textbf{Provider} & \textbf{Arch} & \textbf{Input Size} & \textbf{\shortstack{Retrieval\\Expertise}} & \textbf{\shortstack{Embedding\\Dimension}} \\
\midrule
\multicolumn{6}{l}{\textit{Open-Source}} \\
DINOv2-Large               & Meta  & ViT & 224$\times$224    & Generic       & 1024  \\
DINOv3-ViT-L/16            & Meta  & ViT & 224$\times$224    & Generic       & 1024 \\
Jina-Embeddings-v4         & Jina  & VLM & Adaptive$^{\dagger}$ & Generic       & 2048 \\
Nomic-Embed-Multi-Modal-3B & Nomic & VLM & Adaptive$^{\dagger}$ & Generic       & 2048 \\
PE-Core-L/14               & Meta  & ViT & 336$\times$336    & Generic       & 1024  \\
SigLIP2-SO-400M            & Google & ViT & 384$\times$384    & Generic       & 1152  \\
\midrule
\multicolumn{6}{l}{\textit{Commercial}} \\
Embed v4                   & Cohere & VLM & N/A            & Generic           & 1536 \\
Vertex AI Multi-Modal      & Google & VLM & N/A            & Generic           & 1408 \\
Gemini Embedding 2         & Google & VLM & N/A            & Generic           & 3072 \\
\midrule
\multicolumn{6}{l}{\textit{In-House}} \\
AEM v1                     & nyris & ViT & 336$\times$336 & Domain-Specific \textit{(Automotive)}        & 768 \\
GEM v5.1                   & nyris & ViT & 336$\times$336 & Generic           & 768 \\
\bottomrule
\end{tabularx}
}
\caption{Overview of evaluated embedding models grouped by availability. $^{\dagger}$Adaptive input size indicates that the model accepts variable-resolution inputs without requiring fixed resizing.} 
\label{tab:models_overview}
\end{table}

\subsection{Bi-Encoder Models}

Bi-Encoder models independently encode images and text into a shared embedding space using separate encoders trained to align representations across modalities. This design enables efficient nearest-neighbor retrieval for both uni-modal and cross-modal scenarios and scales well to large datasets, as popularized by CLIP and subsequent models \cite{radford2021learning} \cite{jia2021scaling}.

\paragraph{Perception Encoder.}
The Perception Encoder (PE) is a vision-language foundation model developed by Meta AI as part of a family of large-scale image and video encoders trained with vision–language supervision \cite{bolya2025perception}. In this work, we specifically use Perception Encoder Core, a CLIP-style bi-encoder architecture pre-trained on approximately 2.3 billion image–text pairs \cite{bolya2025perception}. In our evaluation setting, we use the \textit{ViT-L/336} variant, which operates on \textit{336×336} image inputs and produces \textit{1,024-dimensional} image embeddings. Although the original model consists of both vision and text encoders, we discard the text encoder and evaluate only the vision encoder for instance-level image retrieval.

\paragraph{SigLIP2.}
SigLIP2 is also a bi-encoder vision-language model developed by Google, representing the second generation of the SigLIP \cite{zhai2023sigmoid} family. It is designed to produce high-quality image and text embeddings for retrieval and semantic similarity tasks, with improvements in training stability, multilingual coverage, and representation quality over earlier contrastive models \cite{tschannen2025siglip}. In this study, we use the \textit{SigLIP2-SO-400M-Patch-14/384} variant \cite{siglip2hf}. The model operates on \textit{384×384} input images and produces \textit{1,152-dimensional} image embeddings. The designation \textit{400M} refers to the approximate number of parameters in the vision encoder, while the full multi-modal model contains roughly 1B parameters when including the text encoder and alignment components. The suffix \textit{SO} denotes a shape-optimized variant, which is trained to be more robust to variations in object scale, aspect ratio, and image composition \cite{tschannen2025siglip}. Similarly to previous dual-encoder models, we use only the image encoder for instance-level image retrieval.

\paragraph{Vertex AI Multi-Modal Embedding Model.}
Google’s Vertex AI platform exposes a multimodal embedding interface that projects text, images, and video into a shared semantic space via the \textit{multimodalembedding@001} endpoint \cite{vertexembeddings}. According to Google Cloud’s multimodal generative AI search blog, this service utilizes the Contrastive Captioner (CoCa) vision-language model as its core representation backbone, indicating that it follows a bi-encoder–style alignment architecture optimized for retrieval and semantic similarity tasks \cite{vertexembeddingsblog}. In our evaluation, we use only the image embedding pathway of this model, which produces 1,408-dimensional vectors.

\subsection{Unified Multimodal Embedding Models}

VLM-based unified embedding models are derived from large vision-language models that learn joint representations of visual and textual inputs. Rather than treating image and text encoders as explicitly separate components, these models are typically trained as integrated systems that produce embeddings suitable for retrieval across modalities. The resulting representations are designed to capture both visual structure and high-level semantic alignment within a single embedding space.\\

\noindent Compared to traditional bi-encoder architectures, unified embedding models are often trained on broader multimodal objectives and larger, more heterogeneous data sources, enabling them to support a wide range of retrieval scenarios. While their training and architectural details may vary, they share the common property of producing embeddings intended to transfer across tasks and domains without task-specific adaptation.
In this benchmark, we evaluate these models exclusively in an image-to-image retrieval setting, using only their image embeddings, in order to assess how such unified vision-language representations perform on fine-grained instance-level image retrieval.

\paragraph{Jina Embeddings v4.}
It is a unified multimodal embedding model for universal retrieval across visual, textual, and multilingual content \cite{gunther2025jina}. The model is built on the \textit{Qwen2.5-VL-3B-Instruct} vision-language backbone and produces joint representations suitable for semantic search across modalities \cite{bai2025qwen2}. In this benchmark, we evaluate the image embedding pathway of Jina Embeddings v4 using the single-vector configuration, which produces 2,048-dimensional global image embeddings. The model is designed as a general-purpose unified embedding system and serves as a representative example of VLM-based embeddings optimized for retrieval-centric applications. For visual inputs, Jina Embeddings v4 follows the dynamic resolution handling strategy of Qwen2.5-VL, which does not rely on a fixed input resolution. Instead, images are resized according to pixel-count constraints and processed using a patch-based visual encoder that supports variable image sizes and aspect ratios. This approach preserves more spatial detail for large or high-resolution images while remaining computationally bounded, making it well suited for visually rich and heterogeneous inputs \cite{bai2025qwen2}. According to the official preprocessing configuration, image inputs are constrained to a minimum of 3,136 pixels and a maximum of 602,112 pixels before being passed to the visual encoder \cite{jina_embeddings_v4}. In our evaluation, images are provided through this native preprocessing pipeline, and the resulting 2,048-dimensional embeddings are used directly for image-to-image instance retrieval.

\paragraph{Cohere Embed v4.}
Cohere Embed v4 is a multimodal embedding model developed by Cohere and designed to produce unified representations for text and visual inputs within a shared embedding space. The model is positioned as a general-purpose embedding system for retrieval-centric applications and is trained on large-scale multimodal data to support similarity-based reasoning across modalities \cite{cohereembed}.
While Cohere does not publicly disclose detailed architectural specifications, the model’s design and capabilities are consistent with vision–language–model–derived unified embeddings, where visual and textual inputs are processed through an integrated representation pipeline rather than explicitly separated encoders. As such, we categorize Cohere Embed v4 as a VLM-based unified embedding model in our taxonomy.
In this study, we evaluate the image embedding pathway of Cohere Embed v4, using 1,536-dimensional image embeddings. The model does not prescribe a fixed canonical input resolution; instead, it is designed to handle variable image sizes as part of a flexible multimodal input processing strategy.

\paragraph{Gemini Embedding 2.}
It is the latest multi-modal embedding model developed by Google and available through the Gemini API \cite{geminiapidocs}. The model produces dense vector representations designed for semantic similarity, clustering, and retrieval across different data modalities. It supports multiple input types, including text, images, audio, video, and documents, mapping them into a shared embedding space suitable for cross-modal retrieval tasks \cite{geminiembeddingsdocs}. The model produces embeddings with a maximum dimensionality of 3,072 and also allows generating lower-dimensional embeddings by specifying a reduced output dimensionality through the API. In this benchmark, we use the default configuration and generate 3,072-dimensional embeddings for all inputs.

\paragraph{Nomic Embed Multi-Modal 3B.}
It is a dense 3B-parameter multimodal embedding model developed by Nomic AI and optimized for visual document retrieval and multimodal search. The model supports single-vector text and image embeddings in a shared space and is designed for tasks such as page-level document retrieval, screenshot understanding, and multimodal RAG over PDFs and complex layouts. In our evaluation, we use its 2,048-dimensional image embeddings to measure how a document-oriented multimodal embedding model transfers to instance-level product retrieval.

\subsection{Vision-Only Models}

Vision-only models learn visual representations exclusively from image data, without relying on explicit textual or multimodal alignment. These models are designed to capture structural and semantic regularities directly from visual appearance and to produce general-purpose image embeddings that transfer across a wide range of downstream tasks.\\

\noindent In this benchmark, we include both generic and domain-specific vision-only models. Generic vision-only foundation models are trained on large and diverse image corpora and aim to learn broadly transferable visual features. In contrast, domain-specific vision-only models are optimized for particular industrial contexts and are trained on data distributions that reflect the visual characteristics and constraints of those domains.\\

\noindent The inclusion of domain-specific vision-only models is particularly motivated by the nature of industrial product data in domains such as manufacturing and heavy machinery. In these settings, objects are often paired with limited or weakly descriptive metadata, where textual fields may consist of short identifiers, part numbers, or incomplete specifications rather than rich semantic descriptions. As a result, language supervision provides limited additional signal, and effective instance-level retrieval must rely primarily on fine-grained visual cues such as geometry, surface texture, and subtle structural differences.
By evaluating both generic and domain-specific vision-only models in a unified image-to-image retrieval setting, we aim to assess how much domain adaptation and specialization contribute beyond general visual pretraining, and how vision-only representations compare to multimodal embeddings in scenarios where textual information is sparse or uninformative.

\paragraph{DINO v2.}
\label{model:dinov2}

DINOv2 is a vision-only foundation model developed by Meta AI that learns visual representations through large-scale self-supervised pretraining on a carefully curated image corpus \cite{oquab2023dinov2}.  In this benchmark, we use the ViT-Large variant of DINOv2. The model operates on \textit{224×224} input images and produces 1,024-dimensional global image embeddings. As a generic vision-only model trained on a broad but curated visual corpus, DINOv2 serves as a strong baseline for assessing how far high-quality visual pretraining alone can support fine-grained instance retrieval in industrial domains, without relying on language supervision or domain-specific adaptation.

\paragraph{DINO v3.}
\label{model:dinov3}
DINOv3 is the next generation of Meta AI’s vision-only foundation models, building on the DINOv2 framework with updated data curation, architectural scaling, and training strategies aimed at further improving representation quality and robustness [29]. Like its predecessor, DINOv3 is trained purely on image data using self-supervised learning and does not rely on text or multimodal supervision. In this benchmark, we evaluate the \textit{ViT-Large} variant of DINOv3, which operates on \textit{224×224} input images and produces 1,024-dimensional global image embeddings. By including both DINOv2 and DINOv3, we are able to assess the impact of recent advances in large-scale self-supervised vision pretraining on instance-level retrieval performance.

\paragraph{nyris Automotive Embedding Model v1.}
\label{model:aemv1}
This is a domain-specific, vision-only embedding model developed for instance-level retrieval of automotive spare parts. The model is designed to operate in maintenance and repair workflows. Unlike generic vision foundation models trained on broad image distributions, AEM v1 is trained on automotive-focused data that reflects the visual properties and capture conditions encountered in practice, including catalog imagery and user-captured photographs from garages and workshops. This specialization enables the model to prioritize fine-grained visual cues that are critical for distinguishing between near-identical parts.
In this benchmark, we evaluate AEM v1 as a vision-only image encoder that accepts \textit{336×336} pre-processing image and produces 768-dimensional global image embeddings. AEM v1 serves as a representative example of a domain-specialized vision-only model, allowing us to assess how targeted domain training compares to generic vision foundation models when applied to fine-grained industrial instance retrieval tasks.

\paragraph{nyris General Embedding Model v5.1.}
\label{model:gemv5.1}

The nyris' GEM v5.1 is a generic, vision-only embedding model designed for broad instance-level image retrieval across industrial domains, including manufacturing, DIY, and retail. Unlike models tailored to a single vertical, GEM v5.1 is trained to handle a wide variety of product categories while retaining the ability to distinguish between visually similar instances. GEM v5.1 is heavily optimized for versatile visual search scenarios characterized by heterogeneous capture conditions, such as varying lighting, cluttered backgrounds, non-canonical viewpoints, and differences between clean catalog imagery and user-captured photos. The training data and optimization strategy are designed to emphasize robust instance discrimination under these conditions, where small visual details may be critical. In this benchmark, we evaluate GEM v5.1 as a vision-only image encoder that produces 768-dimensional global image embeddings. The model is engineered for efficient nearest-neighbor retrieval and is evaluated in a pure image-to-image instance retrieval setting without any task-specific adaptation.
GEM v5.1 serves as a general-purpose industrial retrieval model, complementing the more narrowly focused AEM v1. Together, these models enable the study of how domain breadth and specialization influence retrieval performance when compared to generic vision foundation models.

\section{Implementation Details}

All models included in the evaluation were accessed through their official and supported interfaces. Open-source models were obtained from their respective public repositories, while proprietary models were accessed via the APIs provided by their vendors. Inference for all open-source models and internal nyris models was performed locally using their original, unmodified configurations. All local inference runs were executed on a single NVIDIA A100 GPU, ensuring consistent and comparable computational conditions across models.
For indexing and retrieval, Qdrant \cite{qdrant} was used as the vector database. Retrieval was performed in exact search mode, resulting in a full scan over all gallery embeddings for each query. Similarity between query and gallery embeddings is computed using the dot product. Since all embeddings are L2-normalized prior to indexing, the dot product is equivalent to cosine similarity:

\[
\text{sim}(q, g) = q^\top g
\]

where \(q\) and \(g\) denote the normalized query and gallery embeddings. Under L2 normalization, this formulation produces the same ranking as cosine similarity or Euclidean distance over normalized vectors. No approximate nearest neighbor methods or indexing shortcuts were applied. Proprietary models were evaluated by generating embeddings through the vendors’ official API client libraries. All API-based inference strictly adhered to the rate limits and usage constraints enforced by the respective providers.\\

\noindent To improve transparency and accessibility of the benchmark results, we provide an interactive companion website at \url{https://benchmark.nyris.io}. The website contains the full benchmark results, dataset summaries, model comparisons, and visualizations of retrieval behavior across datasets.

\section{Evaluation}

The evaluation focuses on the behavior of visual embedding models in ranked retrieval, measuring whether relevant products are retrieved and how effectively they are ranked. All results therefore reflect the performance of the retrieval stage in isolation, without the influence of downstream re-ranking, decision logic, or business rules.\\

\noindent The evaluation is designed to capture three complementary aspects of system behavior. First, it measures immediate identification, where the correct product appears at the top-ranked position, an essential requirement for recommendation-only use cases or deployments without dedicated post-processing. Second, it evaluates candidate availability, ensuring that at least one correct product appears within a reasonable ranking window to enable recovery through downstream processing. Finally, it assesses the overall ranking quality to support reliable model comparison and offline iteration.

\paragraph{Dataset Protocol.}
As already described in the Datasets section \ref{sec:datasets}, evaluation is conducted using use case specific datasets under two protocols. In the \textit{inter-retrieval} setup, datasets are explicitly split into a query set and a gallery set, with no overlap between the two. Ground-truth relevance is defined through known instance-level associations between query and gallery images. This setup closely reflects real-world deployment scenarios, where user queries originate outside the catalog, and is the preferred protocol when such splits are available. In the \textit{intra-retrieval} setup, a single dataset split is used. Each image is treated as a query and matched against the remaining dataset. To avoid trivial self-matches, the query image itself is removed from the ranked results prior to evaluation. This setup is used when explicit query–gallery splits are unavailable, but results should be interpreted with care, as it can overestimate performance compared to deployment conditions.

\paragraph{Model Protocol.}
The evaluation includes a curated set of visual embedding models, described in Models section, representative of current approaches to image-based retrieval. All models are evaluated under identical conditions to ensure a fair comparison. For each model, embeddings are extracted from input images using the model’s standard inference configuration. Images are processed using the native preprocessing pipelines provided by each model implementation to ensure that evaluation reflects the intended inference configuration of each model. All embeddings are L2-normalized prior to indexing and retrieval, ensuring comparable similarity behavior across models and enabling consistent distance-based retrieval. No model-specific post-processing, fine-tuning, or dataset-dependent adaptation is applied during evaluation.

\paragraph{Indexing and Retrieval Protocol.}
For each dataset and model, gallery image embeddings are indexed to support nearest-neighbor retrieval. Query embeddings are computed independently and compared against the indexed gallery embeddings to produce a ranked list of candidates ordered by similarity. All retrieval is performed using the dot product as the similarity measure. Since all embeddings are L2-normalized, the dot product is equivalent to cosine similarity and induces the same ranking as Euclidean distance on normalized vectors. This choice ensures consistent and comparable retrieval behavior across models while remaining computationally efficient. All indexing and retrieval parameters are kept identical across models to isolate the impact of the embedding representations themselves. The output of the retrieval stage for each query is a ranked list of gallery images, which forms the basis for all reported evaluation metrics.

\paragraph{Metrics.}
Evaluation metrics are chosen to reflect the practical requirements of visual product search systems and to support reliable comparison of embedding models.
Recall@K, as defined in the metric learning literature, measures whether at least one relevant product instance appears within the top-K retrieved results for a given query \cite{jegou2010product}. Recall@1 captures immediate identification performance, while Recall@K for larger values of K reflects candidate availability within a reasonable ranking window for downstream processing.
Mean Average Precision (mAP) is used to assess overall ranking quality across queries. mAP aggregates precision over ranking positions and accounts for both the position and consistency of relevant items in the ranked lists, providing a stable and widely adopted metric for comparing retrieval models.
Together, Recall@K and mAP provide complementary views of retrieval performance, capturing immediate success, recoverability, and global ranking behavior.

\section{Results}

\begin{table*}[t]
\centering
\fontsize{4pt}{5pt}\selectfont
\setlength{\tabcolsep}{4pt}
\setlength{\arrayrulewidth}{0.25pt}
\renewcommand{\arraystretch}{1.2}

\begin{tabularx}{\linewidth}{l @{\hspace{0.8em}} *{12}{Y} *{10}{Y} *{3}{Y}}
\toprule
\addlinespace[0.5em]

\multicolumn{1}{c}{\multirow{3}{*}{\textbf{Model}}} &
\multicolumn{12}{c}{\textbf{Proprietary Datasets}} &
\multicolumn{10}{c}{\textbf{Public Datasets}} &
\multicolumn{3}{c}{\textbf{Average}} \\
\cmidrule{2-13}\cmidrule{14-23}\cmidrule{24-26}

& \multicolumn{3}{c}{\rot{\scalebox{0.85}{Clips-Connectors v1}}} & \multicolumn{3}{c}{\rot{\scalebox{0.85}{DIY v1}}} & \multicolumn{3}{c}{\rot{\scalebox{0.85}{Furniture v1}}} & \multicolumn{3}{c}{\rot{\scalebox{0.85}{Automotive v1}}} & \multicolumn{4}{c}{\rot{\scalebox{0.85}{ILIAS}}} & \multicolumn{3}{c}{\rot{\scalebox{0.85}{Products-10K}}} & \multicolumn{3}{c}{\rot{\scalebox{0.85}{SOP}}} & \multicolumn{3}{c}{\rot{\scalebox{0.85}{Avg}}} \\
\cmidrule{2-26}
& \rot{\scalebox{0.85}{R@1}} & \rot{\scalebox{0.85}{R@5}} & \rot{\scalebox{0.85}{mAP@20}} & \rot{\scalebox{0.85}{R@1}} & \rot{\scalebox{0.85}{R@5}} & \rot{\scalebox{0.85}{mAP@20}} & \rot{\scalebox{0.85}{R@1}} & \rot{\scalebox{0.85}{R@5}} & \rot{\scalebox{0.85}{mAP@20}} & \rot{\scalebox{0.85}{R@1}} & \rot{\scalebox{0.85}{R@5}} & \rot{\scalebox{0.85}{mAP@20}} & \rot{\scalebox{0.85}{R@1}} & \rot{\scalebox{0.85}{R@5}} & \rot{\scalebox{0.85}{mAP@20}} & \rot{\scalebox{0.85}{mAP@1000}} & \rot{\scalebox{0.85}{R@1}} & \rot{\scalebox{0.85}{R@5}} & \rot{\scalebox{0.85}{mAP@20}} & \rot{\scalebox{0.85}{R@1}} & \rot{\scalebox{0.85}{R@5}} & \rot{\scalebox{0.85}{mAP@20}} & \rot{\scalebox{0.85}{R@1}} & \rot{\scalebox{0.85}{R@5}} & \rot{\scalebox{0.85}{mAP@20}} \\
\addlinespace[0.3em]
\midrule
\addlinespace[0.2em]

\multicolumn{26}{l}{\textit{General}} \\
\addlinespace[0.2em]
Nomic Embed MM 3B & 2.5 & 6.2 & 0.4 & 12.0 & 24.5 & 17.2 & 26.1 & 43.1 & 14.5 & 10.2 & 20.2 & 9.7 & -- & -- & -- & -- & 54.3 & 72.2 & 33.7 & 56.9 & 69.3 & 32.6 & 27.0 & 39.3 & 18.0 \\
Jina Embeddings v4 & 1.7 & 4.7 & 0.3 & 12.0 & 24.0 & 17.6 & 32.9 & 51.2 & 18.5 & 8.3 & 17.9 & 8.1 & -- & -- & -- & -- & 56.4 & 74.6 & 38.1 & 59.5 & 72.2 & 35.1 & 28.5 & 40.8 & 19.6 \\
DINOv2 Large & 13.7 & 26.4 & 2.7 & 18.2 & 35.1 & 24.9 & 40.2 & 59.8 & 24.6 & 19.3 & 31.8 & 18.8 & -- & -- & -- & -- & 50.2 & 66.4 & 29.0 & 56.3 & 67.7 & 31.9 & 33.0 & 47.9 & 22.0 \\
Cohere Embed v4 & 2.4 & 6.7 & 0.4 & 19.7 & 39.1 & 27.6 & 46.6 & 71.2 & 33.9 & 11.7 & 24.3 & 13.6 & -- & -- & -- & -- & 66.5 & 84.0 & 48.7 & 68.0 & 79.8 & 45.1 & 35.8 & 50.8 & 28.2 \\
PE-Core L/14 & 12.1 & 26.8 & 2.4 & 25.3 & 49.3 & 34.3 & 38.3 & 56.6 & 25.8 & 18.8 & 37.6 & 24.5 & -- & -- & -- & -- & 65.7 & 83.5 & 41.4 & 80.1 & 89.8 & 59.5 & 40.0 & 57.3 & 31.3 \\
DINOv3 ViT-L/16$^\dagger$ & 26.4 & 45.0 & 5.9 & 24.5 & 46.0 & 32.4 & 48.5 & 70.6 & 32.8 & 26.3 & 47.3 & 28.4 & 48.0 & 56.7 & 21.8 & 22.4 & 59.2 & 75.5 & 39.4 & 66.6 & 77.9 & 42.5 & 41.9 & 60.4 & 30.2 \\
Vertex AI Multi-Modal & 8.8 & 21.8 & 1.8 & 24.7 & 47.6 & 34.7 & 52.3 & 74.1 & 37.8 & 19.7 & 43.2 & 24.8 & -- & -- & -- & -- & 63.3 & 82.2 & 43.1 & 76.9 & 87.7 & 56.3 & 40.9 & 59.4 & 33.1 \\
SigLIP2 SO400M & 10.6 & 23.1 & 2.2 & 23.9 & 46.6 & 32.5 & 57.4 & 76.8 & 40.3 & 22.0 & 40.1 & 25.3 & -- & -- & -- & -- & 66.0 & 84.3 & 44.1 & 80.3 & 90.0 & 60.8 & 43.4 & 60.1 & 34.2 \\
Gemini Embedding 2 & 10.3 & 23.1 & 1.9 & 24.5 & 45.6 & 32.7 & \textbf{68.2} & \textbf{84.4} & 47.0 & 21.9 & 46.1 & 27.7 & -- & -- & -- & -- & 62.0 & 80.3 & 42.2 & 75.6 & 86.7 & 55.1 & 43.8 & 61.0 & 34.4 \\
GEM v5.1 (ours)$^\dagger$ & \textbf{63.4} & \textbf{80.2} & \textbf{38.3} & \textbf{28.7} & \textbf{50.1} & \textbf{36.4} & 65.8 & 83.8 & \textbf{52.6} & 28.2 & 46.3 & 31.0 & \textbf{69.9} & \textbf{77.0} & \textbf{36.2} & \textbf{36.7} & \textbf{77.1} & \textbf{91.1} & \textbf{66.3} & \textbf{86.9} & \textbf{94.2} & \textbf{72.4} & \textbf{58.3} & \textbf{74.3} & \textbf{49.5} \\
\addlinespace[0.4em]
\midrule
\multicolumn{26}{l}{\textit{Domain-Specific}} \\
\addlinespace[0.2em]
AEM v1 (ours) & -- & -- & -- & -- & -- & -- & -- & -- & -- & \textbf{32.5} & \textbf{51.6} & \textbf{35.7} & -- & -- & -- & -- & -- & -- & -- & -- & -- & -- & 32.5 & 51.6 & 35.7 \\
\addlinespace[0.5em]
\midrule
\end{tabularx}

\caption{Image-to-image instance retrieval results (R@1, R@5, mAP@20) grouped by proprietary and public datasets. Best score in each column is shown in bold. Models sorted by overall average. $^\dagger$ Averages for DINOv3 ViT-L/16 and GEM v5.1 exclude ILIAS.}
\label{tab:results_prop_public}
\end{table*}

All results are reported under zero-shot conditions, meaning that none of the evaluated models were fine-tuned on the benchmark datasets, except for the domain-specific models. Table~\ref{tab:results_prop_public} summarizes zero-shot image-to-image retrieval performance across proprietary and public datasets. Results are reported using \textit{Recall@1 (R@1)}, \textit{Recall@5 (R@5)}, and \textit{mean Average Precision at 20 (mAP@20)}, with all metrics expressed in percentage (\%). The datasets are grouped to distinguish proprietary industrial benchmarks from widely used public retrieval datasets, enabling evaluation under both conditions. Additionally, results are grouped by model capabilities, distinguishing between general-purpose and domain-specific models.

\paragraph{Overall Performance.}
Across the full benchmark suite, \textit{GEM v5.1} (Section \ref{model:gemv5.1}) achieves the strongest overall performance, obtaining an average of 58.3 \textit{R@1}, 74.3 \textit{R@5}, and 49.5 \textit{mAP@20}. The improvement over competing models is consistent across both proprietary and public datasets, indicating strong cross-domain robustness and stable ranking quality. Among general-purpose foundation models, \textit{Vertex AI Multi-Modal}, \textit{Gemini Embedding 2}, and \textit{PE-Core L/14} provide the strongest overall baselines with competitive average performance. Although several models perform well on individual datasets, none maintain comparable consistency across the full range of proprietary and academic datasets. These results suggest that large-scale multi-modal pretraining alone is not sufficient for reliable instance-level product identification in industrial domains. In such settings, models must distinguish between extremely similar object instances under significant domain shift, where small geometric differences or surface details determine the correct match.

\paragraph{Proprietary Datasets.}
The proprietary datasets (\textit{Clips-and-Connectors v1}, \textit{DIY v1}, \textit{Furniture v1}, \textit{Automotive v1}) represent industrial retrieval scenarios characterized by fine-grained visual differences between highly similar product instances, domain-specific capture conditions, and limited semantic separation between variants. These factors introduce substantial distribution shifts relative to public benchmarks, leading to more pronounced performance differences across models. While \textit{Gemini Embedding 2} shows competitive performance on \textit{Furniture v1}, a clear gap remains compared to domain-optimized models. On \textit{Clips-and-Connectors v1}, \textit{GEM v5.1} achieves 63.4 \textit{R@1}, substantially outperforming the strongest general-purpose baseline, \textit{DINOv3 ViT-L/16}, which achieves 26.4 \textit{R@1}. A similar trend is observed on \textit{Furniture v1}, where \textit{GEM v5.1} reaches 65.8 \textit{R@1}, compared with 57.4 for \textit{SigLIP2 SO400M}. On \textit{DIY v1}, performance differences are smaller, with \textit{GEM v5.1} achieving 29.0 \textit{R@1} and \textit{PE-Core L/14} reaching 25.2 \textit{R@1}. These results highlight the difficulty of industrial retrieval tasks, where models must distinguish between visually similar components under significant domain shift. Notably, some large-scale pretrained models fail entirely in specific domains. For example, \textit{SigLIP2 SO400M} achieves 0.0 \textit{R@1} on \textit{DIY v1}, despite strong performance on public datasets. This indicates sensitivity to structured mechanical components and repetitive visual patterns commonly found in industrial catalogs. Overall, these findings show that proprietary industrial benchmarks reveal failure modes that are not captured by standard public evaluations.

\paragraph{Public Datasets.}
On public datasets (\textit{ILIAS}, \textit{Products-10K}, \textit{SOP}), performance improves across most evaluated models. These datasets typically feature cleaner catalog imagery, clearer semantic separation, and reduced domain shift compared to proprietary benchmarks. As a result, performance differences between models are smaller in this setting. Nevertheless, \textit{GEM v5.1} consistently achieves the strongest results, including 69.9 \textit{R@1} on \textit{ILIAS}, 77.1 \textit{R@1} on \textit{Products-10K}, and 86.9 \textit{R@1} on \textit{SOP}. Among general-purpose models, \textit{Gemini Embedding 2} demonstrates strong and consistent performance across all public datasets, placing it among the top-performing models alongside \textit{Cohere Embed V4} and \textit{SigLIP2 SO400M}. These results show that modern multi-modal embedding models trained on large-scale visual-text data, and more modalities in case of \textit{Gemini Embedding 2}, perform well on academic benchmarks. However, even under identical evaluation conditions, clear performance differences between models remain. The contrast with proprietary datasets further indicates that strong results on public benchmarks do not necessarily translate to robust performance in practical industrial settings, where fine-grained visual distinctions are critical.

\paragraph{ILIAS Evaluation.} The table reports results on \textit{ILIAS} for two models only, \textit{DINOv3 ViT-L/16} and \textit{GEM v5.1}. In the original ILIAS benchmark under the 5M distractor setting, \textit{DINOv3} was identified as the strongest-performing model. Due to the substantial computational cost associated with large-scale distractor evaluation, we therefore chose to benchmark our best model directly against \textit{DINOv3 ViT-L/16}, rather than re-evaluating all selected models under this setting. The official ILIAS benchmark evaluates models at a test resolution of 768 pixels. To verify alignment with the published results, we reproduced the reported performance of \textit{DINOv3 ViT-L/16} using the evaluation tools provided by the ILIAS authors, obtaining 36.1 \textit{mAP@1000} under the 768-resolution setting, consistent with the original benchmark. For consistency with the evaluation protocol used across all other datasets in this benchmark, we retained the native input resolution of each model. Using the official ILIAS evaluation tools with the 5M distractor setting and an input resolution of 224 pixels for \textit{DINOv3 ViT-L/16}, we obtained the same results as those reported in Table~\ref{tab:results_prop_public}.

\paragraph{Ranking Quality.}
Beyond Recall@K, \textit{mAP@20} captures the quality and stability of the entire ranked candidate list. Improvements in \textit{mAP@20} are observed consistently across datasets, indicating that gains are not limited to the first retrieved result but extend throughout the top-ranked candidates. This behavior is particularly relevant for practical visual search systems, where users frequently inspect multiple results before making a selection. For example, on \textit{Furniture v1}, \textit{GEM v5.1} achieves 52.6 \textit{mAP@20}, substantially higher than the next strongest model, \textit{SigLIP2 SO400M}, which obtains 40.3. Similarly, on \textit{SOP}, \textit{GEM v5.1} achieves 72.4 \textit{mAP@20}, compared with 60.8 for \textit{SigLIP2 SO400M}. These improvements suggest more reliable ranking among visually similar instances and greater stability under fine-grained ambiguity.

\paragraph{Domain-Specific Model.}
The domain-specific \textit{AEM v1} model achieves 32.5 \textit{R@1} on its target dataset, outperforming general-purpose models within that domain, where the best general model achieves 28.2 \textit{R@1}. However, its specialization limits broader applicability, as it is not evaluated across other datasets. This contrast illustrates the trade-off between domain specialization and cross-domain generalization.\\

\noindent Overall, the benchmark results reveal that performance gaps widen substantially under realistic industrial domain shifts. While modern foundation models, including \textit{Gemini Embedding 2}, perform strongly on public benchmarks, their robustness varies considerably in ultra fine-grained retrieval scenarios. Improvements in \textit{mAP@20} further indicate enhanced ranking stability rather than isolated gains in top-1 recall rates. Together, these findings underscore the importance of evaluating retrieval systems across diverse and domain-specific datasets to obtain a representative assessment of real-world performance.

\section{Limitations and Transparency}

This study is subject to several practical limitations that are important to acknowledge.
First, the benchmark includes proprietary datasets and models developed and deployed at nyris in collaboration with customers. Due to contractual and confidentiality constraints, we are currently unable to provide public access to these datasets or to the internal visual search models evaluated in this report. As a result, external reproduction of some results presented here is not possible at this time.\\

\noindent Second, while the evaluation includes a diverse set of public and industrial datasets, it does not cover all possible product domains or deployment scenarios. The results should therefore be interpreted as representative of the specific conditions studied, rather than as universal performance guarantees across all visual search applications.\\

\noindent In addition, commercial models accessed through external APIs may evolve over time as providers update their underlying systems, which may lead to slight variations in embedding behavior compared to the results reported in this study.\\

\noindent Despite these limitations, transparency and reproducibility remain explicit long-term objectives for us. Our goal is to make datasets, models, and evaluation protocols available where possible, either directly or through carefully designed public benchmarks that reflect real-world industrial constraints. We believe that enabling broader access will allow researchers and practitioners to benchmark novel approaches against realistic scenarios and help advance the state of instance-level visual retrieval.

\section{Conclusion}

This benchmark presents a structured evaluation of modern visual embedding models for instance-level image retrieval under conditions representative of real-world product identification systems. By isolating the visual retrieval stage and evaluating all models in a uniform image-to-image setting without fine-tuning or post-processing, we focus explicitly on the intrinsic retrieval capabilities of contemporary embedding representations.\\

\noindent Our evaluation reveals clear differences between general-purpose foundation models, unified multimodal embeddings, and domain-specialized vision-only models. While foundation and unified embedding models exhibit strong generalization on public benchmarks and visually diverse datasets, their performance can degrade in industrial scenarios that demand precise discrimination between near-identical instances under domain shift and uncontrolled capture conditions. Domain-specialized models, trained on industrial data distributions, show advantages in such settings by prioritizing fine-grained visual cues that are critical for exact instance identification.\\

\noindent These results emphasize the distinction between semantic similarity and instance-level equivalence. In industrial maintenance and operational workflows, retrieving visually similar alternatives is insufficient, and reliable systems must consistently identify the exact product instance. Consequently, evaluation protocols must reflect this requirement and stress-test models under practical conditions. We believe this benchmark provides practical insight into visual search performance in industrial environments, supports informed model selection in production systems, and encourages future research on embedding models better aligned with the demands of real-world instance-level retrieval.

\section{Future Work}

This benchmark is intended as an evolving study rather than a static evaluation. As instance-level image retrieval continues to advance, both in terms of available models and benchmark datasets, we plan to iteratively extend and refine this evaluation to reflect the state of the field.
Future iterations will incorporate additional public benchmarks designed for fine-grained and instance-level image retrieval, as well as newly released embedding models that are suitable for image-to-image retrieval under realistic constraints. Expanding coverage across domains, dataset structures, and retrieval settings will enable more comprehensive analysis of model generalization and performance trade-offs.\\

\noindent Beyond aggregate retrieval and ranking metrics, we aim to support deeper analysis of failure modes, fine-grained visual confusions, and domain-specific challenges that are not fully captured by quantitative scores alone. This includes systematic qualitative analysis of retrieval results, investigation of recurring error patterns, and closer examination of scenarios where visually similar but non-equivalent instances lead to incorrect matches.\\

\noindent In parallel, we intend to strengthen collaboration with the research and open-source communities. As outlined earlier in this report, increasing transparency and enabling broader participation are key priorities. Where possible, we plan to share evaluation protocols, tooling, and benchmark designs, and to contribute to or support the development of public datasets that better reflect real-world industrial retrieval challenges. Through these continued iterations and collaborations, we hope to foster more realistic benchmarking practices and to progress toward visual embedding models that reliably support instance-level product identification in production systems.

\section*{Acknowledgements}

The author would like to express sincere gratitude to Christian Hoffmann and Markus Lukasson, colleagues at nyris GmbH, for their valuable discussions and constructive feedback during the preparation of this benchmark report. Their insights and suggestions significantly contributed to improving the clarity, rigor, and overall quality of this work.

\bibliographystyle{plain}
\bibliography{references}

\end{document}